\documentclass[runningheads]{llncs}
\usepackage[T1]{fontenc}
\usepackage{multirow}
\usepackage{graphicx,verbatim}
\usepackage{amsmath}
\usepackage{graphicx} 
\usepackage{hyperref}

\begin{document}
\title{EndoLRMGS: Complete Endoscopic Scene Reconstruction combining Large Reconstruction Modelling and Gaussian Splatting}
\titlerunning{EndoLRMGS: Endoscopic Scene Reconstruction}  

\author{Xu Wang\inst{1,2,5} \and Shuai Zhang\inst{1,3} \and Baoru Huang\inst{4} \and Danail Stoyanov\inst{1,3} \and Evangelos B. Mazomenos\inst{1,2}}  

\authorrunning{X. Wang et al.}  

\institute{
UCL Hawkes Institute, University College London, UK \and
Dept of Medical Physics \& Biomedical Engineering, University College London, UK \and
Dept of Computer Science, University College London, UK \and
Dept of Computer Science, University of Liverpool, UK \and
\email{xu.wang.23@ucl.ac.uk}
}

\maketitle              
\begin{abstract}
Complete reconstruction of surgical scenes is crucial for robot-assisted surgery (RAS). Deep depth estimation is promising but existing works struggle with depth discontinuities, resulting in noisy predictions at object boundaries and do not achieve complete reconstruction omitting occluded surfaces. 
To address these issues we propose \textbf{EndoLRMGS}, that combines Large Reconstruction Modelling (LRM) and Gaussian Splatting (GS), for complete surgical scene reconstruction. GS reconstructs deformable tissues and LRM generates 3D models for surgical tools while position and scale are subsequently optimized by introducing orthogonal perspective joint projection optimization (OPjPO) to enhance accuracy. In experiments on four surgical videos from three public datasets, our method improves the Intersection-over-union (IoU) of tool 3D models in 2D projections by>40\%. Additionally, EndoLRMGS improves the PSNR of the tools projection from 3.82\% to 11.07\%. Tissue rendering quality also improves, with PSNR increasing from 0.46\% to 49.87\%, and SSIM from 1.53\% to 29.21\% across all test videos. 
\footnote{Code is available at: https://github.com/MichaelWangGo/EndoLRMGS}


\end{abstract}
\section{Introduction}

\begin{figure}[!t]
\includegraphics[width=1.0\textwidth]{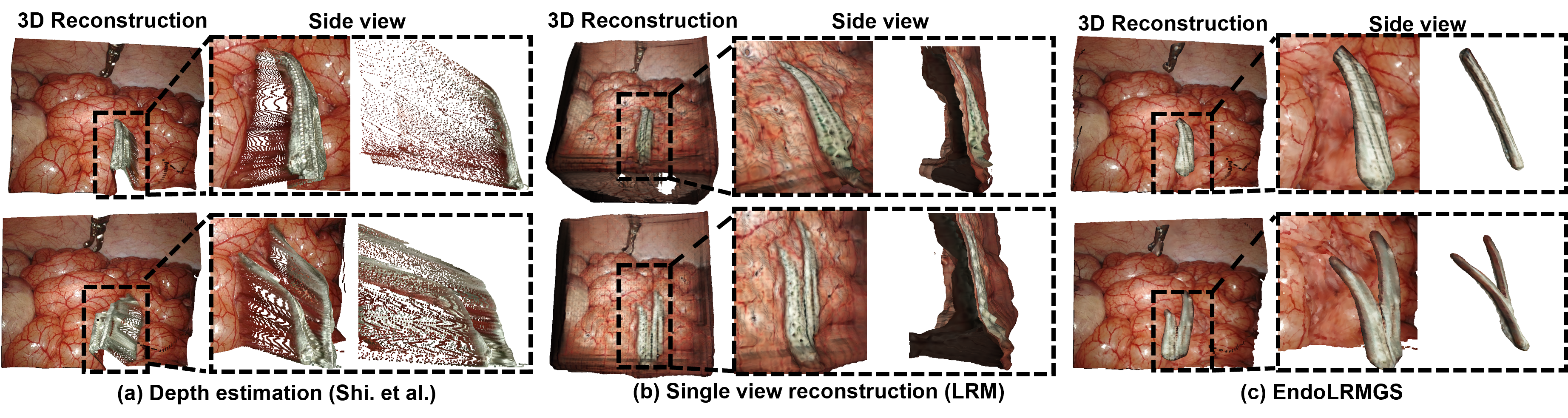}
\caption{Reconstruction results from depth estimation~\cite{shiBidirectionalSemiSupervisedDualBranch2023}, single view reconstruction~\cite{hongLRMLARGERECONSTRUCTION2024}, and the proposed EndoLRMGS. EndoLRMGS reconstructs a complete and clear 3D surgical scene, including occluded tissue and closed surgical tools. Side views reveal that noisy points that should not appear in a faithful reconstruction, are eliminated.}
\label{fig:figure1}
\end{figure}

Reconstruction of surgical scenes from endoscopic videos is a crucial yet challenging task in robot-assisted surgery (RAS) \cite{xu2024review}. It enhances surgical precision \cite{yang2024efficient} and supports navigation with augmented/virtual reality \cite{zhang2022autostereoscopic} and RAS automation \cite{chadebecq2023artificial}. Depth estimation, an important step for scene reconstruction, typically relies on feature matching and multi-view geometry \cite{shiBidirectionalSemiSupervisedDualBranch2023,bae2022multi} and this in endoscopic videos is susceptible to errors at depth discontinuities in boundary regions (e.g. 2D projection junctions between surgical tools and tissues). Moreover, only the depth of visible surfaces can be obtained, neglecting information in occluded areas, like tissues behind tools and the back side of the tools themselves, resulting in incomplete 3D reconstruction. As shown in Fig.~\ref{fig:figure1}(a), noisy points appear at the edges of the tools and the occluded regions behind them are not reconstructed.

Recent efforts explore neural radiance fields (NeRF) and differentiable splatting for scene rendering and reconstruction. EndoNeRF \cite{wang2022neural} uses mask-guided ray casting to handle tools occlusion and reconstruct deformable tissue, while EndoSurf \cite{zha2023endosurf} employs a signed distance field (SDF) to improve quality. Gaussian Splatting (GS) \cite{kerbl20233d} leverages anisotropic 3D Gaussians and renders images with a tile-based rasterizer. Both NeRF and GS synthesize views based on observed camera viewpoints, limiting their ability to reconstruct unobserved surfaces, such as the back sides of surgical tools. Despite promising results single-object 3D reconstruction, methods such as BundleSDF \cite{wenBundleSDFNeural6DoF2023} and the Large Reconstruction Model (LRM) \cite{hongLRMLARGERECONSTRUCTION2024}, which learns 3D priors from one million samples, can faithfully generate 3D models for rigid objects but struggle with deformable surfaces. Additionally, they lack spatial awareness of the relationship between tools and background tissue (see Fig.~\ref{fig:figure1}(b)). These limitations make them unsuitable for RAS tool reconstruction, as surgical instruments consist of multiple articulated parts, making it difficult to accurately estimate their poses and reconstruct them as a single object. 

In this paper, we propose \textbf{EndoLRMGS}, a novel approach leveraging LRM and GS for complete endoscopic scene reconstruction. Our contributions are: 1) Surgical tools are treated as closed objects, while anatomical tissue is considered as a surface object, enabling tailored reconstruction strategies for each; 2) Tools are identified using zero-shot decoupled video segmentation and tracking  DEVA~\cite{chengTrackingAnythingDecoupled2023}, and an initial 3D model is predicted from a single input image using LRM. Then, GS is employed to render RGB and depth information of occluded areas, facilitating the reconstruction of the tissue surface. 3) Orthogonal Perspective Joint Projection Optimization (\textbf{OPjPO}) is introduced to address scale and spatial positioning inconsistencies among the the generated point clouds. Our work is the first to achieve complete reconstruction of both tissues and surgical tools in endoscopic scenes, effectively handling depth discontinuities while preserving the geometric integrity of 3D objects (see Fig.~\ref{fig:figure1}(c)).

\section{Method}

\begin{figure}[!b]
\label{fig:pipeline}
\centering
 \includegraphics[width=1.0\textwidth]{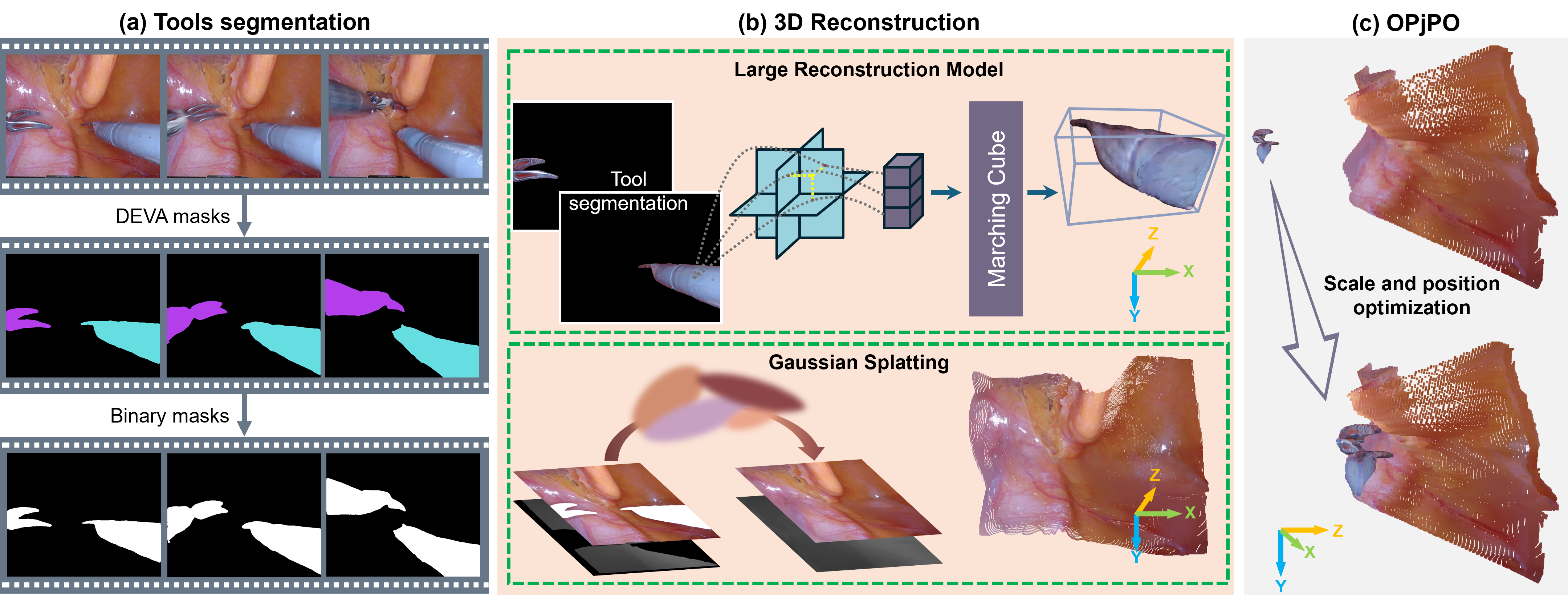}
\caption{The EndoLRMGS framework includes surgical tool segmentation with unique mask assignment, separate reconstruction of surgical tools and tissue, and the solution of scale and position uncertainties through the proposed OPjPO method. }
\end{figure}

The EndoLRMGS architecture is illustrated in Fig.~\ref{fig:pipeline} and consists of three components: 1) tool segmentation; 2) 3D reconstruction of  tissue (surface-object) and tools (closed-object) and 3) position/scale optimization by proposed OPjPO. The output is a watertight complete 3D model of the endoscopic scene.

\subsubsection{Segmentation:}
For differentiating the surgical tools from the background tissue, each of which requires a tailored reconstruction strategy, we use DEVA\cite{chengTrackingAnythingDecoupled2023} to produce a unique mask for each tool across all frames, as shown in Fig.~\ref{fig:pipeline}(a). This enables to determine the scale factor and optimize the position of each tool. The segmentation is also used for mask-guided rendering, to reconstruct the occluded anatomy tissue. 

\subsubsection{Full surgical scene reconstruction:}
We propose a novel approach for endoscopic scene reconstruction (see Fig.~\ref{fig:pipeline}(b)) consisting of two key components: EndoGaussian for tissue (surface-object) and LRM for surgical tools (closed-object) reconstruction. We utilize a single RGB image $I_i$, depth map $D_i$, and binary mask $M_i$ that delineates surgical tools. Following \cite{liuEndoGaussianRealtimeGaussian2024}, we back-project each frame into a point cloud $P_i$ and merge them for holistic initialization:
 \begin{equation}
 \begin{aligned}
P = \{ P_1, P_2, \dots, P_N \},
\quad
P_i = K^{-1} T_i D_i (I_i \odot M_i)
\label{equation:initilization}
\end{aligned}
\end{equation}
$K$ and $T_i$ are known camera intrinsic and extrinsic parameters and $\odot$ denotes element-wise product. Then the scene can be initialized to Gaussian primitives. The color $\hat{ C} (x) $and depth $\hat{D}(x) $ of a pixel $x$ can be rendered as:
 \begin{equation}
 \begin{aligned}
\hat{C} (x)= \sum_{i=1}^{n} c_i \alpha_i \prod_{j=1}^{i-1} (1 - \alpha_j),
\quad
\hat{D}(x) = \sum_{i=1}^{n} d_i \alpha_i \prod_{j=1}^{i-1} (1 - \alpha_j),
\end{aligned}
\end{equation}
where $c_i$ and $d_i$ are the color and depth derived from the spherical harmonic coefficients of $i$-th Gaussian, and $\alpha_i$ is determined a 2D covariance matrix multiplied by the opacity. 
To enforce consistency between predicted and real colors and depths. The color loss is defined as:  
\begin{equation}
\mathcal{L}_{\text{color}} = \sum_{x \in I} \left\lVert M(x) \left( \hat{C}(x) - C(x) \right) \right\rVert_1,
\end{equation}  
where \(M(x)\) is the binary tool mask, and \(C(x)\) represents the color in input image at pixel \(x\). 
For depth consistency, we define the following loss function:  
\begin{equation}
\mathcal{L}_{\text{depth}} = \sum_{x \in I} \left\lVert M(x) \left( \hat{D}^{-1}(x) - D^{-1}(x) \right) \right\rVert_1 + 1 - \frac{\text{Cov}(M \odot \hat{D}, M \odot D)}{\sqrt{\text{Var}(M \odot \hat{D}) \text{Var}(M \odot D)}},
\end{equation}  
The first term ensures consistency in inverse depth predictions, while the second term enforces structural alignment between predicted and real depths by maximizing their correlation. \(\text{Cov}(\cdot, \cdot)\) is the covariance and \(\text{Var}(\cdot)\) the variance.  
In parallel, each surgical tool is processed individually by LRM. The tool segmentation is first converted into patch-wise feature tokens using DINOv2 \cite{caron2021emerging}, which are projected into a triplane representation. An MLP layer subsequently predicts RGB color and density for each triplane point within a 3D grid sampled at a resolution of 384×384×384. Finally, the marching cubes algorithm extracts a complete mesh from the rendered density field, yielding detailed initial 3D models of the tools. Reconstructed tools within this fixed grid do not have physically meaningful dimensions and accurate spatial positioning. 

\subsubsection{Orthogonal perspective joint projection optimization:}
\begin{figure}[!t]
        \centering 
        \includegraphics[width=1.0\textwidth]{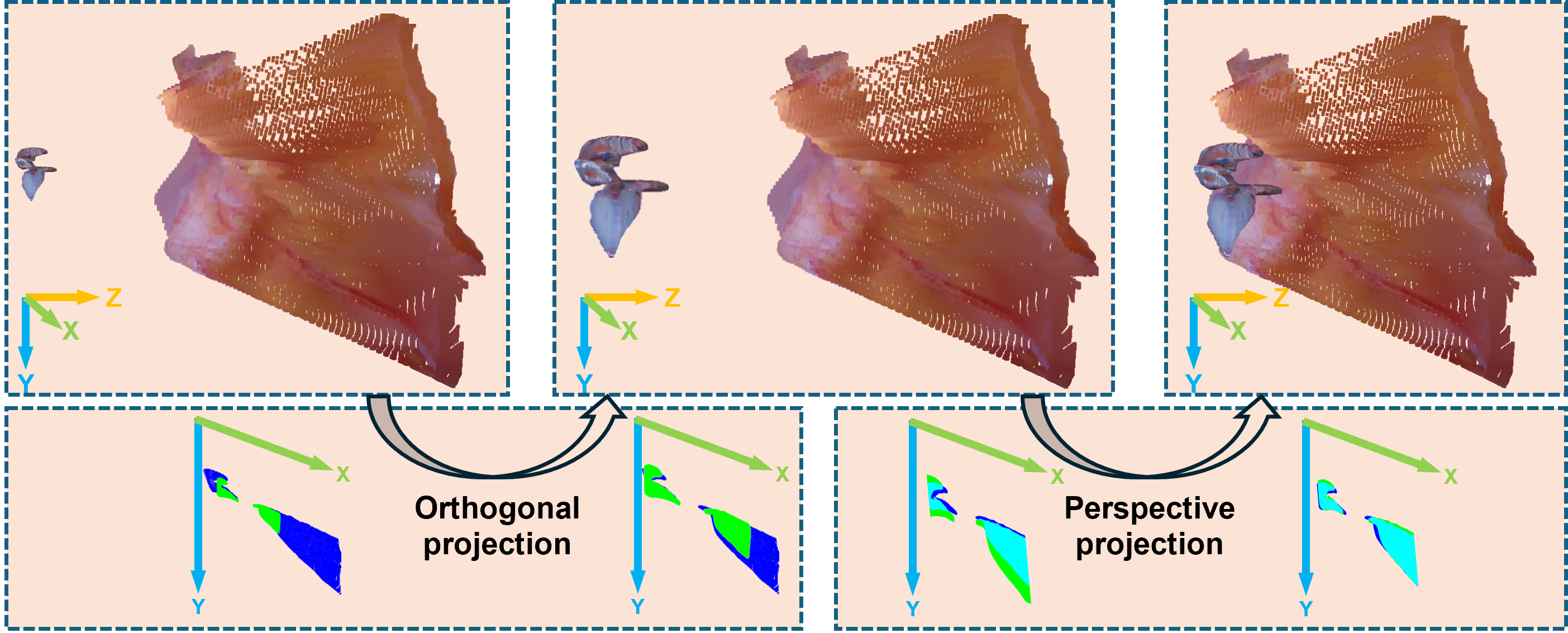}
        \caption{The OPjPO framework for solving scale factor and position uncertainty. The scale factor of a tool 3D model is first determined by orthogonal projection, and then its spatial position is determined by perspective projection.} 
\label{fig:optimization}
\end{figure}

To address this, we exploit that LRM-generated 3D models align with the coordinate axes of the GS reconstruction and design an OPjPO method, as illustrated in Fig.~\ref{fig:optimization}. Given the narrow surgical workspace, where the distance between tools and tissue is minimal, the 3D mapping of 2D tool masks onto the tissue closely approximates their true size. This enables to first estimate the tool's scale and then refine its spatial position. We use the GS-reconstructed tissue point cloud, which is composed of $Q$ 3D points: $ P = \{ p_i = (x_i, y_i, z_i) | i = 1,2,\dots,Q \} $,  to construct an orthogonal transformation matrix $M_o$:
\begin{equation}
M_o = \begin{bmatrix} 
S & \mathbf{0} \\ 
\mathbf{t}^\top & 1 
\end{bmatrix},\ 
S=\operatorname{diag}\!\left(\frac{2}{r-l},\, \frac{2}{t-b},\, \frac{-2}{f-n}\right), 
\mathbf{t} = 
\begin{bmatrix}
-\frac{r+l}{r-l}, 
-\frac{t+b}{t-b}, 
-\frac{f+n}{f-n}
\end{bmatrix}^\top,
\end{equation}
where $l=\min_{i} (x_i)$,  $r=\max_{i} (x_i)$ ,  $b=\min_{i} (y_i)$,  $t=\max_{i} (y_i)$,  $n=\min_{i} (z_i)$,  $f=\max_{i} (z_i)$. Then, 
the tool point cloud $P_{tool}$ and masked tissue point cloud $P_{tissue[mask]}$ are projected onto a 2D space using $M_o$ without perspective distortion:
 \begin{equation}
 \begin{aligned}
\hat{I}_{tool}^{i}=M_{o}\cdot P_{tool}^{i},\
\hat{I}_{tissue[mask]}^{i}=M_{o} \cdot P^{i}_{tissue[mask]},
\label{equation:orthogonal}
\end{aligned}
\end{equation}
where $i$ represents the index of the tool. Finally, the scale factor is solved by:
 \begin{equation}
\sigma = \sqrt{A_{mask} / A_{tool}}, 
\end{equation}
where $A_{mask}$ and $A_{tool}$ are the bounding box areas of the orthogonal projections $\hat{I}_{tissue[mask]}$ and $\hat{I}_{tool}$, respectively.

Next, the tool point cloud is scaled by $\sigma$ and projected by perspective projection matrix $M_p = K[R|T]$, where $K$ represents the camera intrinsics, $R$ is the identity matrix and $T$ is the zero vector:
\begin{equation}
 \begin{aligned}
\tilde{I}_{tool}^{i}=\sigma \cdot M_{p} \cdot P_{tool}^{i}, \
\tilde{I}_{tissue[mask]}^{i}=M_{p} \cdot P^{i}_{tissue[mask]}.
\end{aligned}
\end{equation}
Finally, the spatial position of each tool is optimized by adjusting its coordinates along the $x$, $y$, and $z$ axes to maximize the intersection-over-union (IoU) between the tool projection and the actual tool mask. This ensures accurate alignment between the reconstructed tool and its real-world counterpart.

\section{Experiments}
\subsection{Datasets and implementation}
We evaluate our method on four videos that contain camera/tool motion and deformable tissue, from three publicly available datasets: StereoMIS \cite{hayoz2023pose} provides 1280 × 1024  videos from procedures with a da Vinci Xi (Intuitive Surgical Inc., USA) surgical robot in three in-vivo porcine subjects. We select a clip from video P2\_6. EndoNeRF \cite{wang2022neural} provides two 640 × 512 videos from da Vinci robotic prostatectomy. We select clips from both. SCARED \cite{allan2021stereo} contains 1280 × 1024 da Vinci Xi videos of porcine cadaver anatomies. We use dataset\_6, keyframe\_4 sequence. All datasets provide camera calibration parameters. Our experiment directly employs the inference capabilities of DEVA\cite{chengTrackingAnythingDecoupled2023} for segmentation and LRM\cite{hongLRMLARGERECONSTRUCTION2024} for tools 3D model generation.
EndoGaussian\cite{liuEndoGaussianRealtimeGaussian2024} is trained on each of the four videos following the reoccomended 7:1 train/test split and with tool masks generated from DEVA and refined by a dilation operation with a 47×47 kernel. Experiments are coded in PyTorch and run on an NVIDIA RTX 4090. We assess reconstruction accuracy with photometric error metrics, PSNR, SSIM, and LPIPS, and evaluate tool scale and position using IoU.

\subsection{Qualitative Results}
\begin{figure}[!t]
        \centering
        \includegraphics[width=1.0\textwidth]{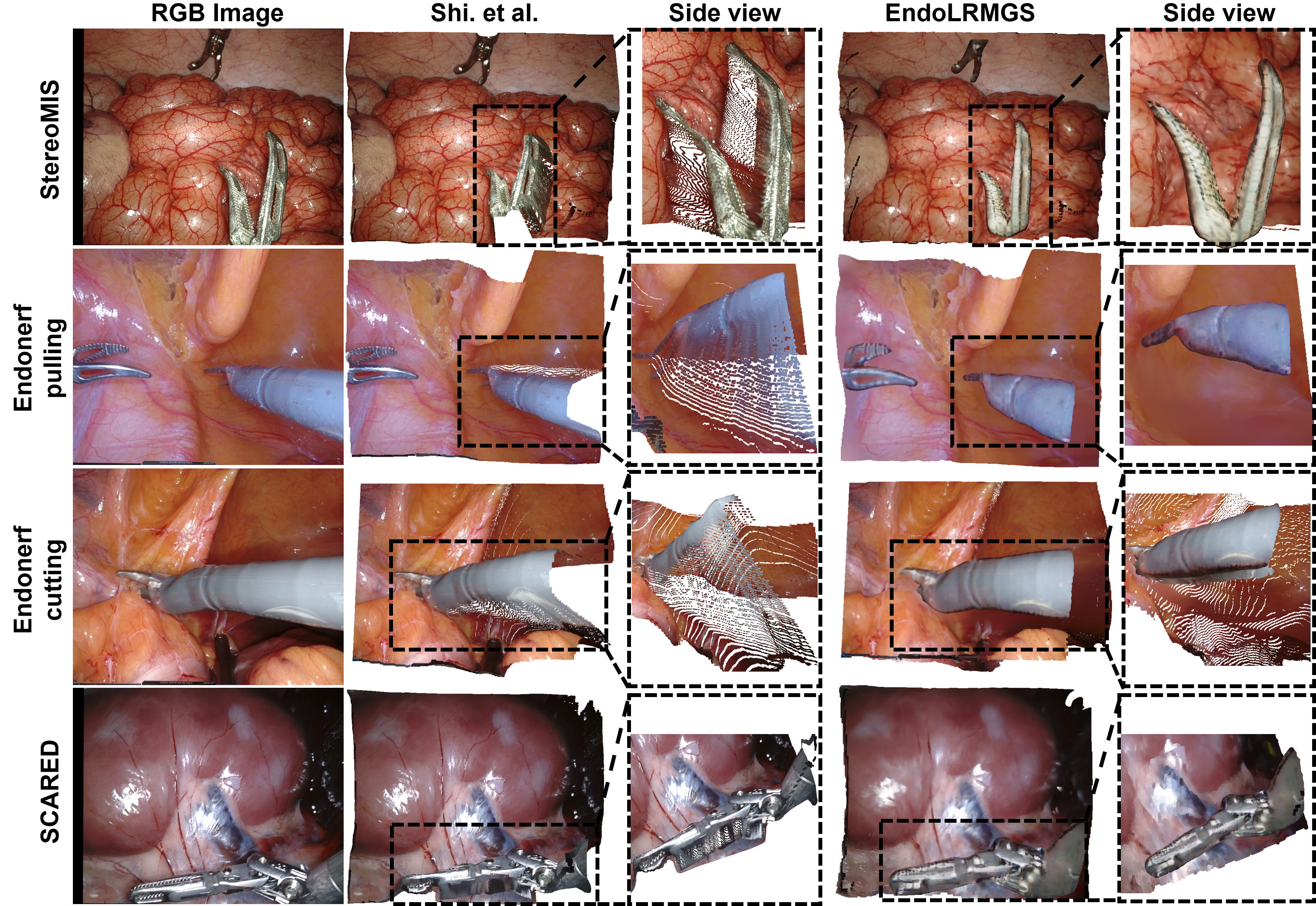}
\caption{3D Reconstruction of Shi. et al. and EndoLRMGS in different situations: deformable tissue and moving surgical tools (StereoMIS and EndoNerf) and moving camera (SCARED).} 
\label{fig:visualization}
\end{figure}

We qualitatively evaluate EndoLRMGS against reconstruction via conventional depth estimation from Shi. \textit{et al.} \cite{shiBidirectionalSemiSupervisedDualBranch2023}, pretrained on Scene Flow and SCARED. Outputs are shown in Fig.~\ref{fig:visualization} and we observe that the conventional approach results in incomplete 3D models and reconstructs only the visible surfaces of surgical tools. Evidently, EndoLRMGS renders both the occluded RGB image and depth information, reconstructing the full geometry of the scene, including occluded regions. Our method also effectively mitigates point cloud noise at the edges of tools caused by depth discontinuities  (see Fig.~\ref{fig:visualization} side views). \textbf{The full point clouds are provided in the supplementary video.}

\subsection{Quantitative Results}
Projecting the entire 3D model onto a 2D image can cause multiple 3D points to map to the same pixel, leading to evaluation inaccuracies. Thus, we separately project the surgical tools and tissue onto 2D images and compare each projection with the corresponding region in the input image. Depth estimation methods, which reconstruct directly from RGB images, are excluded from evaluation. 
\begin{figure}[!t]
\centering
\includegraphics[width=1.0\textwidth]{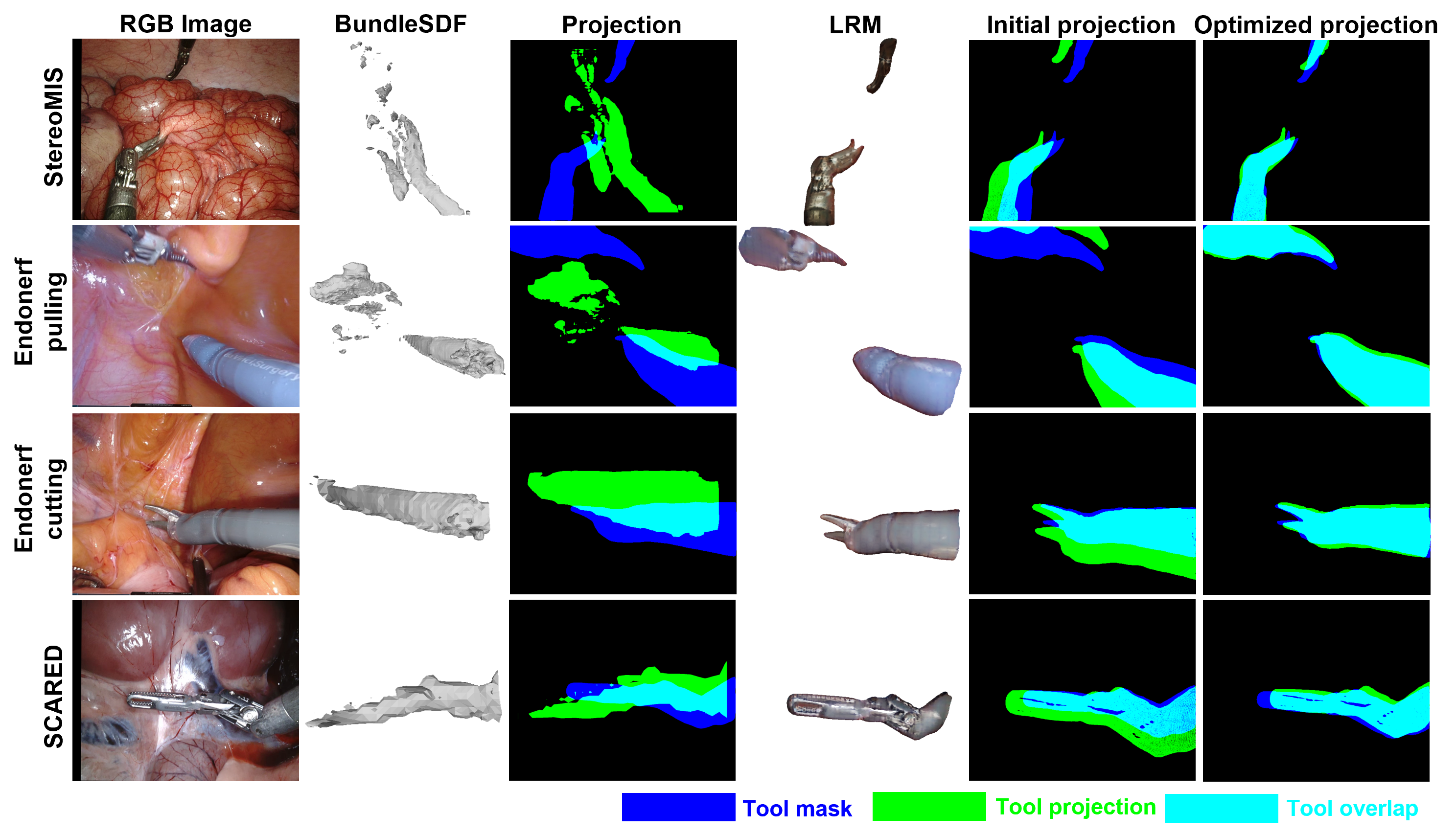}
\caption{Comparison of 3D surgical tool reconstruction and 2D projections across three datasets. Columns: (1) RGB images, (2) BundleSDF 3D reconstruction, (3) BundleSDF projection with ground truth masks, (4) LRM 3D reconstruction, (5) LRM projection with ground truth masks, (6) Optimized projection using our method.} 
\label{fig:projection}
\end{figure}
\subsubsection{Tools reconstruction: } 
We compare EndoLRMGS with two state-of-the-art (SOTA) object reconstruction methods, BundleSDF and LRM. BundleSDF is trained with tools segmentation provided by DEVA~\cite{chengTrackingAnythingDecoupled2023} and depth maps provided by Shi. \textit{et al}. ~\cite{shiBidirectionalSemiSupervisedDualBranch2023}. For LRM, we directly use a well-trained model on Objaverse~\cite{deitkeObjaverseUniverseAnnotated2023} and MVImgNet~\cite{yu2023mvimgnet} to generate 3D models from the tool segmentations. Per-frame results are generated from the four surgical videos.

Fig.~\ref{fig:projection} highlights the limitations of existing methods in reconstructing surgical tools. BundleSDF, which relies on object pose estimation to accumulate depth information, struggles with articulated parts in a tool, leading to inaccurate reconstructions (second and third columns). LRM can generate high-quality 3D models but it lacks spatial awareness, resulting in misalignment with real-world scale and position(fourth and fifth columns). In contrast, proposed OPjPO effectively optimizes the scale factor and position of LRM-generated 3D models, producing projections that closely align with ground truth masks (final column). 

From Table.~\ref{tab:instruments}, EndoLRMGS yields superior tools reconstruction results against SOTA methods. On average, EndoLRMGS yields an IoU of 81.23\%, significantly surpassing LRM (41.12\%) and BundleSDF (18.65\%). To assess appearance similarity, we compare 2D projections of the reconstructed models with original images within the regions covered by tool masks. Our method outperforms both, with average improvement of 1.2990 in PSNR, 0.0062 in SSIM, and a reduction of 0.0176 in LPIPS. While our results are close to those of LRM, EndoLRMGS further refines the size and spatial position of the 3D model using OPjPO, ensuring better consistency with real-world scale and positioning. 




\begin{table}[!t]
\centering
\caption{Quantitative metrics of surgical tools and tissue.}
\label{tab:instruments}
\scalebox{0.8}{
\begin{tabular}{l |l |l |l |l |l |l |l |l |l} \hline 
 & \multicolumn{5}{c|}{Tools assessment}& \multicolumn{4}{c}{Tissue assessment}\\ \hline    
 Datasets & Methods &  IoU&PSNR & SSIM & LPIPS  & Methods & PSNR & SSIM &LPIPS \\ \hline
\multirow{3}{*}{\shortstack[c]{StereoMIS \\ (P2\_6)}}& BundleSDF &   2.90&7.7077&  0.1316&  0.9427 & Endo4DGS& 26.6333& 0.7550&0.3612\\ 
 & LRM &  21.57 &20.2489 & 0.9364 & 0.1071  & Endosurf
& 20.6426& 0.5052&\textbf{0.2840}\\ 
 & EndoLRMGS&  \textbf{79.49}&\textbf{21.1444} & \textbf{0.9389}& \textbf{0.0943} & EndoLRMGS& \textbf{27.8582}& \textbf{0.7985}&0.2985\\ \hline 
\multirow{3}{*}{\shortstack[c]{Endonerf\\pulling}} & BundleSDF &   7.94&6.7422&  0.2667&  0.7050 & Endo4DGS& 25.4823& 0.9216&0.1296\\  
 & LRM &  46.87 &15.3594 & 0.8178 & 0.2822  & Endosurf
& 28.0909& 0.9304&0.1216\\  
 & EndoLRMGS&  \textbf{83.44}&\textbf{17.2711}& \textbf{0.8302}& \textbf{0.2463} & EndoLRMGS& \textbf{28.2214}& \textbf{0.9449}&\textbf{0.1071}\\ \hline
\multirow{3}{*}{\shortstack[c]{Endonerf \\ cutting}} & BundleSDF &   25.70&5.6127&  0.5652&  0.7279 & Endo4DGS& 19.8647& 0.9009&0.1240\\  
 & LRM &  62.50 &17.6953 & 0.8464 & 0.2421  & 
Endosurf
& 26.7815& 0.9194&0.1200\\ 
 & EndoLRMGS&  \textbf{85.83}&\textbf{19.2483}& \textbf{0.8550}& \textbf{0.2185} & EndoLRMGS& \textbf{27.2964}& \textbf{0.9392}&\textbf{0.0928}\\ \hline
\multirow{3}{*}{\shortstack[c]{SCARED\\(d6k4)}}& BundleSDF & 38.05&7.6487&  0.1720&  0.7839 & 
Endo4DGS& 12.9647& 0.6091&0.5770\\  
 & LRM &  33.52 &18.8214 & 0.9197 & \textbf{0.1162} & Endosurf
& 13.8160& 0.4738&0.5808\\  
 & EndoLRMGS&  \textbf{76.16}&\textbf{19.5693}& \textbf{0.9209}& 0.1181  & EndoLRMGS& \textbf{27.5606}& \textbf{0.8605}&\textbf{0.3228}\\ \hline \hline
\multirow{3}{*}{Average}& BundleSDF & 18.65& 6.9278& 0.2839& 0.7899& Endo4DGS& 21.2363& 0.7967&0.2980\\
 & LRM & 41.12& 18.0313& 0.8801& 0.1869& Endosurf
& 22.3328& 0.7072&0.2766\\
 & EndoLRMGS& \textbf{81.23}& \textbf{19.3083}& \textbf{0.8863}& \textbf{0.1693}& EndoLRMGS& \textbf{27.7342}& \textbf{0.8858}&\textbf{0.2053}\\ \hline
\end{tabular}}
\end{table}

\subsubsection{Tissue reconstruction:} 
We compare EndoLRMGS tissue reconstruction with two SOTA methods, Endo-4DGS~\cite{huang2024endo} and EndoSurf~\cite{zha2023endosurf}. Each video's frames are split 7:1 for training and testing excluding regions covered by tool masks.
Results are listed in Table~\ref{tab:instruments}, with EndoLRMGS showing superior performance against the SOTA across all datasets. Especially for the case with significant camera motion (SCARED), EndoLRMGS surpasses Endo-4DGS and EndoSurf by 14.5959 and 13.7446 in PSNR. On average, our approach improves PSNR by 5.4014, SSIM by 0.0891, and reduces LPIPS by 0.0713 compared to other two SOTA methods.

\section{Conclusion}

This paper presents EndoLRMGS, a novel reconstruction method that integrates LRM and GS to generate complete 3d models of surgical scenes. We adopt EndoGaussian to reconstruct deformable tissue and utilize LRM to generate a initial tool 3D model. We then resolve scale factors, determining object dimensions and refine the spatial positioning of the tool 3D model by introducing an OPjPO method without requiring a reference point cloud. Results on surgical videos presenting different challenges demonstrate that our method achieves SOTA in both tool reconstruction (average IoU of 81.23\%, PSNR of 19.3083, SSIM of 0.8863, and LPIPS of 0.1693) and tissue reconstruction (average PSNR of 27.7342, SSIM of 0.8858, and LPIPS of 0.2053). These results highlight the effectiveness of our approach in  reconstructing complete surgical scenes with high accuracy.

    



%
%
%
\bibliographystyle{splncs04}
\bibliography{Myreferences}
%




\end{document}